\documentclass[
twocolumn,
]{ceurart}

\usepackage{listings}
\usepackage{tabularx}

\begin{document}

\copyrightyear{2023}
\copyrightclause{Copyright for this paper by its authors.
  Use permitted under Creative Commons License Attribution 4.0
  International (CC BY 4.0).}

\conference{Workshop on Artificial Intelligence for Access to Justice (AI4AJ 2023), June 19, 2023, Braga, Portugal.}

\title{LLMediator: GPT-4 Assisted Online Dispute Resolution}

\author[1]{Hannes Westermann}[%
orcid=0000-0002-4527-7316,
email=hannes.westermann@umontreal.ca,
]
\cormark[1]
\address[1]{Cyberjustice Laboratory, Faculté de droit, Université de Montréal, Québec, Canada}

\author[2]{Jaromir Savelka}[%
orcid=0000-0002-3674-5456,
email=jsavelka@cs.cmu.edu,
]

\address[2]{School of Computer Science, Carnegie Mellon University, Pittsburgh, USA}

\author[1]{Karim Benyekhlef}[%
orcid=0000-0001-9390-556X,
email=karim.benyekhlef@umontreal.ca,
]

\cortext[1]{Corresponding author.}

\begin{abstract}
In this article, we introduce LLMediator, an experimental platform designed to enhance online dispute resolution (ODR) by utilizing capabilities of state-of-the-art large language models (LLMs) such as GPT-4. In the context of high-volume, low-intensity legal disputes, alternative dispute resolution methods such as negotiation and mediation offer accessible and cooperative solutions for laypeople. These approaches can be carried out online on ODR platforms. LLMediator aims to improve the efficacy of such processes by leveraging GPT-4 to reformulate user messages, draft mediator responses, and potentially autonomously engage in the discussions. We present and discuss several features of LLMediator and conduct initial qualitative evaluations, demonstrating the potential for LLMs to support ODR and facilitate amicable settlements. The initial proof of concept is promising and opens up avenues for further research in AI-assisted negotiation and mediation.
\end{abstract}


\maketitle

\section{Introduction}
Many individuals face difficulties while resolving their legal disputes. This is especially the case in areas of high-volume, low-intensity disputes (such as debt, consumer and employment issues), which often affect laypeople, i.e., individuals without legal training \cite{farrow2016everyday}. Laypeople may often not understand what legal rules apply to their situation, which can be an impediment to enforcing their rights \cite{currie2009legal}. Even if they are aware of their rights, they may struggle to come to a resolution (see, e.g., \cite{savage2022experiences}). Bringing the dispute to court could entail significant monetary, temporal and psychological costs \cite{semple2015cost}. Unresolved legal issues can be very stressful and they may even impact individuals' health. Unresolved legal issues were estimated to cost society 746 million dollars in Canada alone in 2014 \cite{farrow2016everyday}. Globally, the United Nations Task Force on Justice estimates that 1.5 billion individuals are unable to resolve their legal issues at any point in time \cite{justice_for_all}. 

A popular way of increasing access to justice is the use of so-called alternative dispute resolution (ADR) methods, i.e., methods of solving disputes outside of the traditional judicial system \cite{carneiro2014online}. Such methods include, e.g., negotiation or mediation. During negotiation, the parties communicate to each other, attempting to reach an amicable agreement. 
In mediation, a neutral third party is expected to establish a rapport with the parties, to bring creativity in thinking about potential solutions, and to identify overlapping interests between the parties that may be helpful in achieving a consensus between them \cite{erichson2019dark, cole2006online}.
These processes can  be carried out online, in so-called online dispute resolution (ODR) \cite{benyekhlef2005online}. This may have several advantages, including being more efficient and convenient for the individuals, who do not need to physically attend a court. The cooperative and non-adversarial nature of these approaches to dispute resolution may help the parties devise an amicable solution that both parties feel ownership of \cite{carneiro2014online,benyekhlef2005online,branting2022computational,schmitz2019measuring}. ODR can be a very powerful approach---e.g., in PARLe-OPC, an ODR platform developed by the Cyberjustice Laboratory and implemented with the Quebec consumer protection authority, 70\% of cases are settled, 45\% at the stage of negotiation and 25\% at the stage of mediation \cite{benyekhlef2020intersection}.

To foster further improvements in the usability and efficacy of ODR systems, we present LLMediator---an experimental platform focused on exploring how large language models (LLMs) such as GPT-4 may be used to support the parties and the mediator in arriving at a settlement. LLMs have recently risen to fame as very powerful AI systems. Models such as GPT-4 \cite{openai2023gpt4} or LLama \cite{touvron2023llama} are able to perform various tasks based on a textual prompt, which is often not too far from instructions understandable to a human. The models can also be accessed via an application programming interface (API), which allows developers to integrate the capabilities into their own applications.

LLMediator is an experimental system leveraging GPT-4 to enhance an ODR platform. The system can reformulate users' messages to be, e.g., less emotional and more conducive to reaching a settlement. Further, LLMediator can suggest draft messages for the mediator (or facilitator), taking into account previous messages by the parties and special instructions by the mediator. Finally, LLMediator offers an experimental feature to autonomously respond to the discussion by the parties. In this paper, we demonstrate selected features of LLMediator and perform initial qualitative evaluations of this system.

\section{Related Work}
Using artificial intelligence (AI) to increase access to justice is an often studied question. For example, systems have been built that seek to help users understand their rights in certain situations (see, e.g., \cite{branting2001advisory,thompson2015creating,bickel2015online,paquin1991loge,westermann2023justicebot}). In this work, we present an approach to facilitate dispute resolution via augmented intelligence tools for negotiation and mediation.

Integrating AI into the negotiation process has previously been explored by multiple authors. One possible such integration is the display of a so-called ``Best Alternative to the Negotiated Agreement'' (BATNA) during negotiation, which can provide contextual information helping the users find a fair agreement (see e.g. \cite{carneiro2014online,zeleznikow2016can,susskind2019online,benyekhlef2018intelligence,bickel2015online}). Other approaches include using game-theory methods to elicit an acceptable settlement (see, e.g., \cite{bellucci2005developing,lodder2003role}), and by structuring the dialogue between parties \cite{kistemaker2021rechtwijzer}. Here, we detect messages by the parties containing inflammatory language, and suggest alternative ways of conveying the message, that may be more amenable to amicable resolution.

Similarly, AI systems have been built to support mediation. Branting et al. developed a system that monitors messages exchanged during negotiation for situations that require an immediate human intervention, such as conflict escalation or inactivity. The system can further suggest standard messages, that may be relevant to a particular dispute, to the facilitator \cite{branting2022computational}. Researchers have further investigated the possibility of using ChatGPT to suggest relevant questions and potential party priorities to to support the mediator in their role \cite{Bergman_2023}. In this work, we evaluate the ability of LLMs to generate suggestions for the mediator, or potentially even intervene in a negotiation autonomously, which could support mediators or even be used to provide certain mediation services in areas where traditional mediation is not feasible (e.g., due to the costs).

Huang and Ouyang proposed a technique for developing a judge perspective generation model in the context of ODR. The approach employs pre-trained LLMs, and is adaptable to various scenarios through minimal low-code fine-tuning \cite{huang2022generate}. 

Our work explores the use of LLM models, e.g., GPT-4, to support legal tasks. There has been a growing interest in exploring capabilities of GPT models in such applications. Yu et al. applied GPT-3 to the COLIEE legal entailment task that is based on the Japanese Bar exam, substantially improving over the existing state-of-the-art \cite{https://doi.org/10.48550/arxiv.2212.01326}. Similarly, Bommarito II and Katz utilized GPT-3.5 for the Multistate Bar Examination \cite{bommarito2022gpt}.
The model performed surprisingly well but not well enough to pass the exam. Later, Katz et al. applied GPT-4 to the entire Uniform Bar Examination (UBE) and observed the system passing the exam \cite{katz2023gpt}. The use of GPT models has further been explored in the context of the assessment of trademark distinctiveness \cite{goodhue2023classification}, legal reasoning \cite{blair2023can,nguyen2023well}, U.S. Supreme court judgment modeling \cite{hamilton2023blind} and giving legal information \cite{tan2023}. Likewise, GPT-4 has been used to annotate textual legal data based on annotation guidelines \cite{savelka2023GPT4,savelka2023unlocking} and to generate explanations of legal concepts based on sentences from statutes \cite{savelka2023concepts}. Here, we use the models to increase access to justice, by supporting parties in negotiation and mediation.

In this work, the key uses we explore employ GPT-4 as an augmented intelligence tool, that supports the users in their decision-making as opposed to directly providing instructions or advice, which can decrease the risk associated with such tools. Previously, this approach has been explored in the legal domain to generate boolean search rules \cite{westermann2019computer}, perform annotations more efficiently \cite{villata2020sentence}, and provide legal information to laypeople \cite{westermann2023justicebot, westermann2023bridging}.

\section{Example use cases}
\label{sec:example}

\begin{figure}[t] 
    \centering
    \includegraphics[width=7cm]{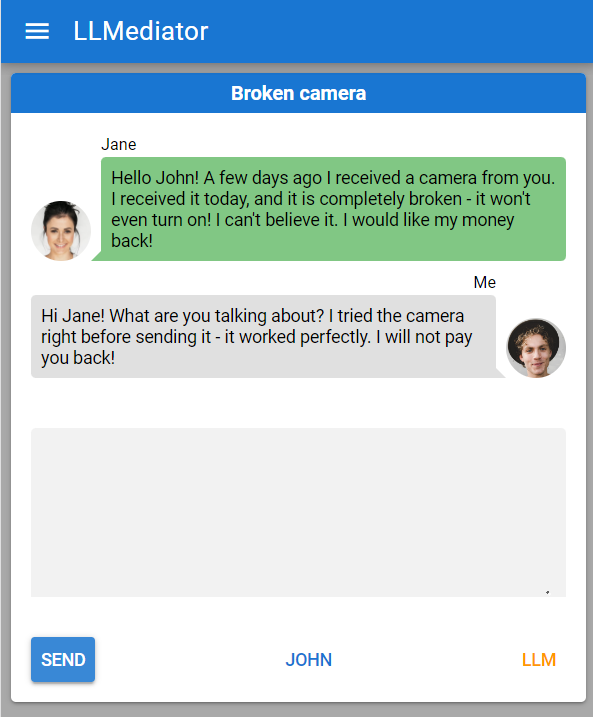}
    \caption{A screenshot of the LLMediator interface, showing a dispute regarding a broken camera.}
    \label{fig:screen01}
\end{figure}

LLMediator is a web-based application that contains the necessary features to explore and evaluate the integration of LLMs into the negotiation and mediation processes. Upon entering the system, the user is exposed to a chat interface, that allows them to discuss their issue with the other party, with the aim to come to an amicable solution, as can be seen in Figure \ref{fig:screen01}. In the experimental prototype, the user can select whether they wish to intervene as one of the parties (John or Jane) or the mediator. This part of the platform is similar to other existing ODR platforms that allow the parties to communicate in a convenient and asynchronous manner without having to physically relocate.

However, the LLMediator platform also has several novel aspects, that use the power of LLMs to support various tasks in the mediation platform. These features include:
\begin{enumerate}
    \item F1 - Suggesting reformulations of inflammatory messages to be less emotional and, hence, more conducive to finding an amicable settlement
    \item F2 - Drafting message suggestions for mediators to more efficiently intervene in a discussion
    \item F3 - Potentially, autonomously intervening in a discussion to guide the parties towards an amicable solution
\end{enumerate}

\noindent The technical implementation of these features is described in Section \ref{sec:tech}.

\subsection{F1 - Reformulating inflammatory messages}
\begin{figure}[t] 
    \centering
    \includegraphics[width=7cm]{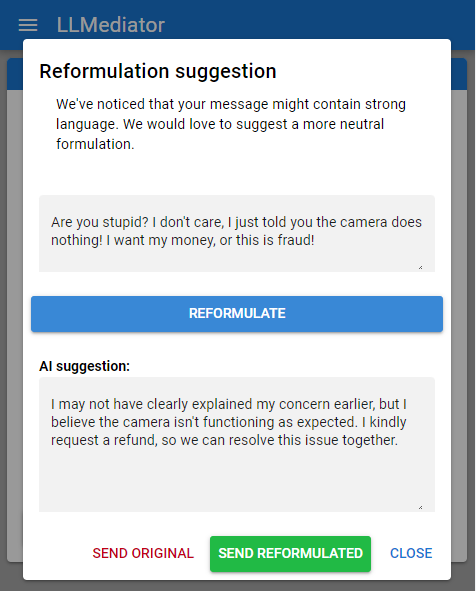}
    \caption{The LLMediator has detected a message that is not helpful for an amicable settlement. It suggests a reformulation that the user may consider to send instead.}
    \label{fig:screen02}
\end{figure}

Emotions can run high in negotiation situations. In Figure \ref{fig:screen02}, the user intends to send a message that is inflammatory, and may derail attempts to find a mutually acceptable solution. The LLMediator platform detects this language, and offers an alternative formulation to the user, generated by GPT-4. The user can choose to send the original message, send the reformulated message directly, or edit and then send the reformulated message. The suggestion may nudge the user toward stating their request in a more neutral fashion, which could be more helpful for arriving at an amicable resolution to their dispute.

\subsection{F2 - Drafting messages for the mediator}
When negotiation is in deadlock or does not result in a settlement, it may be useful for a neutral, third-party mediator to intervene in the discussion. Let us consider the role of the mediator to see how the LLMediator can support them in the task of encouraging the parties to arrive to a friendly resolution.

\begin{figure}[t] 
    \centering
    \includegraphics[width=7cm]{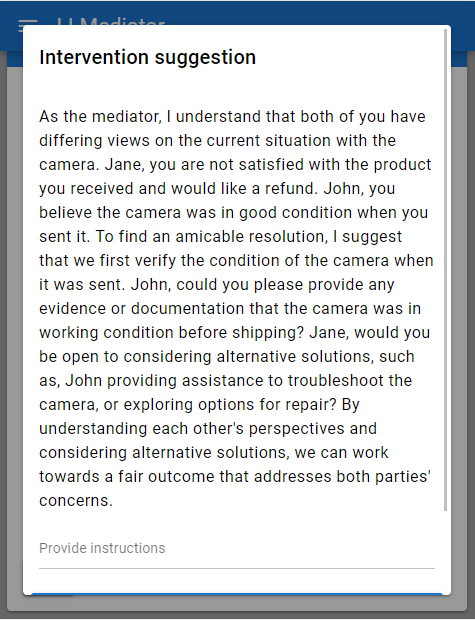}
    \caption{The LLMediator suggests a possible intervention for the mediator.}
    \label{fig:screen03}
\end{figure}

The mediator can press a button in the interface to generate a draft intervention. This feature uses GPT-4 to read the previously sent messages and draft a suggested message that gently guides the parties toward a friendly resolution. Figure \ref{fig:screen03} shows such a suggested intervention.

\begin{figure}[t] 
    \centering
    \includegraphics[width=7cm]{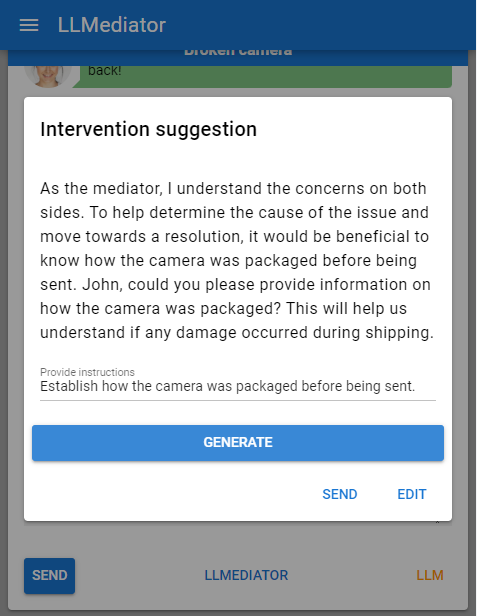}
    \caption{The LLMediator incorporates specific instructions by the mediator in generating a message.}
    \label{fig:screen04}
\end{figure}

This message may serve as a starting point for the mediator to decide how to intervene. They have the option to send the message, or edit it and then send it. Further, they can provide additional instructions to GPT-4, guiding the model to emphasize certain aspects. For example, the mediator may guide the parties to consider the packaging of the camera. Figure \ref{fig:screen04} shows a screenshot of what happens when the mediator instructs the model to explore this issue with the parties.

\subsection{F3 - Autonomously intervening in the negotiation?}

In some situations, it may make sense to allow the model to automatically intervene in the negotiation. This could be the case when the dispute value is too low for employing a human mediator, or if there are not enough mediators to attend to all disputes in a certain area. 

\begin{figure}[t] 
    \centering
    \includegraphics[width=7cm]{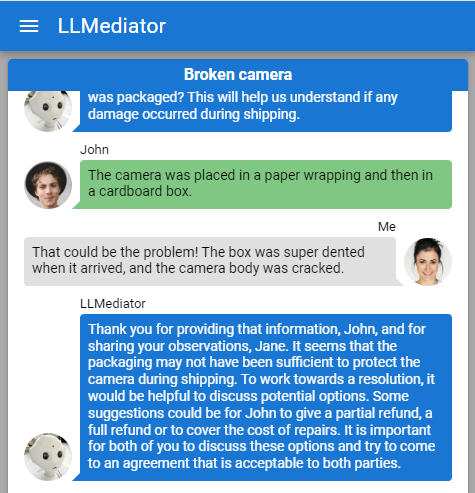}
    \caption{The LLMediator could potentially automatically intervene in a discussion.}
    \label{fig:screen05}
\end{figure}

In Figure \ref{fig:screen05}, one of the parties requested an intervention by the LLMediator, to move the discussion forward. The LLMediator autonomously generated a message and sent it to the parties, suggesting a few possible options to encourage a settlement. There may also be other triggers to move the system to intervene, such as a period of inactivity, or inflammatory messages being sent (see Section \ref{sec:triggers}). As will be discussed in Section \ref{sec:discussion:F3}, this approach may carry with it certain risks that would need to be carefulyl studied before it could be deployed.

We have now seen the three current capabilities of the LLMediator to support the parties in a mediation. As we can see, the system suggests relevant and context-adapted reformulations and interventions, that may encourage the parties to use calmer language, or save mediator's time. Next, let us take a look at the technical considerations behind the system.

\section{Technical considerations}
\label{sec:tech}
For the features described in the preceding section, a number of choices need to be considered. These involve questions such as when the features should be activated, how the LLM should be prompted to achieve the best results, and what to do with the resulting text. In this section, we will explore these considerations.

\subsection{Large language model used}
For the current version of the system, we used GPT-4 (Generative Pre-trained Transformer). GPT-4 is a multi-modal pre-trained LLM created by OpenAI \cite{openai2023gpt4}. The model has shown impressive performance on a wide variety of tasks, including passing the Uniform Bar Examination \cite{katz2023gpt}. Some researchers even argue that the model shows sparks of general intelligence \cite{bubeck2023sparks}. We interacted with the 8k token version of the model via the OpenAI API.\footnote{GitHub: OpenAI Python Library. Available at: \url{https://github.com/openai/openai-python} [Accessed 2023-02-09]}

\subsection{F1 - Reformulating inflammatory messages}
The first capability of the LLMediator is the reformulating of inflammatory messages. The motivation behind this feature is that individuals may have a lot of emotions attached to their dispute, which could be reflected in the tone of the messages. Such messages may increase the tension and aggravation between the parties rather than decrease it, making it less likely that the parties will find an amicable agreement, which is the goal of the negotiation process.

In order to overcome the issue, the LLMediator contains a feature that aims to reformulate the message of the user to a more neutral tone, while preserving the overall semantic content of the message. This way, the parties are able to communicate their opinions, without escalating the tone of the conversation.

\subsubsection{Detect a message requiring intervention}
\label{sec:detecting}
In order to only intervene when necessary, the LLMediator needs a way to detect whether a message contains inflammatory language and would benefit from reformulation. Thus, the first step is to classify the tone of the message. It should be noted that the level of emotions and anger in a message that should lead to an intervention is a policy decision---different thresholds may lead to different results. Further, one could imagine multiple methods of detecting whether a message may be inflammatory. These methods include:
\begin{enumerate}
    \item Contacting an LLM, such as GPT-4, with every message and inquiring whether it is inflammatory or not. Depending on the volume of messages on a platform, this method may be expensive and introduce latency into the platform, which may be confusing for the user, since the message needs to be analyzed before it is sent to the other party.
    \item Training a bespoke machine learning model to detect whether a message warrants intervention. This approach was used by the authors in \cite{branting2022computational}. The researchers trained a model on previous messages from an ODR court, to detect whether an immediate intervention was necessary, and suggest standard message responses.
    \item Using an off-the-shelf machine learning model. Sentiment detection is a well-studied field, with many available models. These models could be used to detect messages with a negative polarity, or even to detect certain emotions (such as angry or sad), and use thresholds to trigger an intervention.
    \item Using keyword search. The simplest approach in detecting inflammatory words is simply to scan the message sent by the user for inflammatory words, such as ``a**hole''. 
    \item The user may decide that they would like a suggestion to reformulate a message, e.g., if they are aware that their message may come across as angry or defensive, and would like a suggestion as to how they may express themselves in a more neutral fashion.
\end{enumerate}

For the LLMediator platform, the main focus is investigating whether the model is able to reformulate the message in a useful way. Therefore, we chose options 4 and 5. The system scans the message for inflammatory terms, and suggests a reformulation if any such terms are found. Further, the user has the option to manually activate the reformulation dialog. More sophisticated methods to detect inflammatory messages will be investigated in future work.

\subsubsection{Reformulating the message}
Once the inflammatory message has been detected, the next step is to prepare a reformulated message with a more constructive tone, while maintaining the core meaning of the message. The message is thus sent to an LLM, together with a prompt instructing the LLM to reformulate the message. For the LLMediator, we are using the GPT-4 API provided by OpenAI, together with the following prompt:

\textit{``You are an ODR platform. You are given the chat message of a party. Reformulate the message to maintain the content, but make it less confrontational and more encouraging for an amicable settlement. Respond directly with the reformulated message, do not explain.''}

\subsubsection{Showing the reformulation to the user}
\label{sec:show_reformulation}
Once the inflammatory message has been detected and reformulated, it needs to be re-introduced into the system in a useful manner. Ways of doing this include: 

\begin{enumerate}
    \item The reformulated message could be sent directly to the other party.
    \item The user could be provided with the reformulated version of their message, and given the option to send or edit the reformulated message, or send their original message.
\end{enumerate}

For the LLMediator, we chose the second option, since this preserves the control of the user over their communication, while potentially nudging them toward a more positive tone. The user is thus shown the reformulated message, and has the option to either accept and send the reformulated version, edit the reformulated version and then send it, or to send the original message.

\subsubsection{Additional examples}
Table \ref{table:messages} shows a few additional examples of reformulated messages:

\begin{table}[htbp]
  \centering
  \begin{tabularx}{\linewidth}{|X|X|}
    \hline
    \textbf{Original message} & \textbf{Reformulated message} \\
    \hline
    What the ****? I told you about the water leak weeks ago and you did nothing! Fix it or I will see you in court! & I wanted to remind you that I brought up the water leak issue a few weeks ago. It would be great if we could find a solution to address it before considering legal action. \\
    \hline
    You still have not repaid me the 1000 USD I lent you! You are the worst friend ever, we are done! & It seems that the 1000 USD I lent you hasn't been repaid yet. As friends, let's discuss this issue and work towards resolving it amicably. \\
    \hline
    Here is what happened: I told you that the tree was hanging over my lawn many on the 3rd of April. On the 15th, it was still there, so I cut it down. This is your ******* fault, you could have fixed it!! & I noticed on April 3rd that the tree was overhanging my lawn. Despite addressing the issue, it remained unchanged by the 15th, which led me to cut it down. I believe this situation could have been avoided if timely action was taken on your part. \\
    \hline
  \end{tabularx}
  \caption{Example of original and reformulated messages}
  \label{table:messages}
\end{table}


\subsection{F2 - Drafting messages for the mediator}
Another feature of the LLMediator is to provide a suggestion for an intervention that a mediator might want to send to the parties of a dispute. In mediation, a third-party mediator aims to intervene in a discussion to guide it toward a friendly settlement between the parties. Such an intervention can be very powerful and positive.

In the LLMediator, we provide a feature that allows the mediator to receive a suggestion for a message that they may consider as an intervention in a discussion by the parties. The suggestion of pre-defined messages that a mediator may want to use, based on a discussion, were previously explored in \cite{branting2022computational}. In our research, the suggested messages are generated ad hoc using an LLM, in our case the GPT-4 API provided by OpenAI. The mediator can launch the draft feature by clicking a button in their chat interface.

\subsubsection{Generating the message suggestion}
\label{sec:message_generation}

In order to generate the message, we use the following prompt:

\textit{``
You are a mediator. Your goal is to guide the discussion of two parties towards an amicable settlement that is acceptable to both parties. Respond to this communication between the parties. Stick to the role of the mediator - do not complete the dialog of the parties. Remain neutral, do not take the side of any party.
''}

This prompt was arrived at after empirical investigations. Further, we provide the model with a context of the 10 latest messages in the conversation. In case the mediator enters an additional instruction (see figure \ref{fig:screen04}, these instructions are also sent to the model. Based on this context, the model can gain an understanding of the discussion and hopefully provide a helpful draft message to the mediator.

\subsubsection{Further examples}
Table \ref{table:suggestions} shows a number of examples where the situation described above in section \ref{sec:example} was used, after an additional follow-up by John inquiring why he should be liable when the postal service is clearly at fault. The table presents the suggested intervention by the model, optionally together with additional instructions provided by the mediator.

\begin{table*}[htbp]
  \centering
  \begin{tabularx}{\linewidth}{|X|X|}
    \hline
    \textbf{Mediator instructions} & \textbf{Suggested intervention} \\
    \hline
    None & Thank you for expressing your concern, John. It's clear that the situation isn't ideal for either party. One possible solution could be to file a claim with the postal service to seek reimbursement for the damaged camera. That way, Jane can receive compensation for the broken camera and the responsibility would shift to the postal service. Would both of you be open to trying this approach to reach a resolution? \\
    \hline
    Inquire whether there might be an insurance offered by the trading platform used & John, I understand your concern. It might be possible that the trading platform you have used for the transaction offers some form of insurance or buyer/seller protection. In order to consider this as an option, could you please let us know which platform you used for the transaction and if they offer anything in this regard? This might help both of you reach a fair and amicable resolution. \\
    \hline
    Ask the parties to clarify the model, value and state of the sold good. & I understand your concerns, John. However, it's important to consider that part of the responsibility lies in the packaging of the item to ensure its safe delivery. In order to evaluate the options more fairly, could both of you please provide more information about the camera, such as the model and the estimated value, as well as its condition at the time of the sale? This will allow us to further discuss the possible solutions mentioned earlier and find a resolution that both parties find satisfactory. \\
    \hline
    
  \end{tabularx}
  \caption{Example of suggested interventions, based on the dispute around the broken camera and optional instructions provided by the mediator.}
  \label{table:suggestions}
\end{table*}



\subsection{F3 - Autonomously intervening in the negotiation?}

The most radical use of the LLMediator is the direct intervention of the system in the negotiation. Here, the system acts as a mediator itself. While this could be a very interesting and powerful use-case, it also poses a number of substantial risks, as will be explored in section \ref{sec:discussion:F3}. Thus, considerable studies would have to be performed prior to the deployment of such a system. Here, it is presented as purely experimental feature.

\subsubsection{Triggers}
\label{sec:triggers}

For the LLMediator to be able to intervene, there needs to be a trigger to determine when it should send a messages. Some possibilities for such a trigger include:

\begin{enumerate}
    \item Intervening when there has been a period of inactivity (compare \cite{branting2022computational})
    \item Intervening when the discussions between the parties are getting heated (compare \cite{branting2022computational} and section \ref{sec:detecting}).
    \item Intervening every few messages (e.g., every 10 messages sent) to shape and guide the discussion.
    \item Intervening when asked to by either of the parties.
\end{enumerate}

Since the focus of the LLMediator is to evaluate the capability of the LLM models to interact with the user in a useful way, we have currently only implemented the last option.

\subsubsection{Generating the message}
Once it has been determined that the system should intervene, the previous messages are sent to the LLM, which will generate a response that can be provided by the system to the user. In order to generate the message, the same procedure as described above in section \ref{sec:message_generation} is used. However, in this case, the message is not given to the human mediator, but rather sent directly to the parties in the conversation.

\subsubsection{Additional examples}

To better understand the types of messages suggested by the LLMediator, table \ref{table:intervention} lists a few example disputes, and the messages generated by the system.

\begin{table*}[htbp]
  \centering
  \begin{tabularx}{\linewidth}{|X|X|}
    \hline
    \textbf{Dispute summary} & \textbf{Generated intervention} \\
    \hline
    A tenant (Jane) wants specific performance and damages for a water leak that they supposedly informed their landlord (John) of 5 weeks ago, claiming the landlord did not respond. The landlord claims that they were never informed. & As a mediator, I would like to help Jane and John resolve this issue. It appears there may be a misunderstanding about the communication taken place. Firstly, let's try to establish the facts. Jane, could you please provide more information about when and how you informed John about the water leak? And John, is there any possibility that you might have missed or overlooked this communication? Let's work together to find a fair and acceptable solution for both parties. \\
    \hline
    A person (John) wants damages for lost wages, after falling on a road in front of the other parties apartment. The other party (Jane) claims that they did clear the snow that day. & As your mediator, I understand that both of you have concerns and perspectives on this issue. John, you experienced an accident that resulted in lost wages due to the snow outside Jane's apartment. Jane, you claim that you had cleared the snow earlier that day. To move forward, let's first establish the extent of responsibility each party has in this situation. This includes discussing the circumstances of the accident further and any relevant information, such as local laws or regulations regarding snow removal. Would you both be willing to discuss in more detail the specifics of the incident and the snow removal practices at Jane's apartment? This way we can better understand the situation and work towards an amicable agreement. \\
    \hline
    John borrowed 400 CAD from Jane to repay some credit card debt, but became unemployed and is now unable to repay. Jane wants the money back with interest. Multiple messages have been exchanged back and forth. & Thank you for providing more context about the situation. It seems like the initial agreement was informal and based on John's promise to repay when he got his next paycheck. However, John is currently unemployed, which makes the repayment more challenging. As a suggestion, would both of you be open to discussing a repayment plan that takes John's current financial situation into consideration without burdening Jane? This could include re-assessing the interest or agreeing on a feasible timeframe to repay the debt. \\
    \hline
    Jane ordered seeds from Johns website, but never received them. John claims to habe no record of the order. A heated conversation ensues. & As the mediator in this situation, I would like to remind both parties to remain respectful during this discussion. 
    Jane, I understand that you have concerns regarding the status of your order, and John, I hear that you have no record of the transaction. Let's try to work together to identify and resolve the issue. Jane, would you mind providing any evidence or details regarding your order, such as a transaction ID, order number, or a confirmation email? This will help John to verify your order in his system. 
    John, please be patient while we gather this information, and once we have it, I kindly ask you to look into your system to confirm Jane's order. 
    By acting in a respectful and cooperative manner we can work on finding a solution that satisfies both parties. \\
    \hline
    
  \end{tabularx}
  \caption{Example of generated interventions based on a different disputes.}
  \label{table:intervention}
\end{table*}





\section{Discussion}
In this paper, we have demonstrated selected capabilities of LLMediator, an experimental platform to explore the use of LLMs in an ODR context. The platform currently offers several features, including (F1) the reformulation of messages to a more neutral tone, (F2) the assisting of the mediators by providing suggested interventions, and (F3) the optional direct intervention of the system in the negotiation. Let us discuss selected aspects of these different features, including their potential and possible risks.

\subsection{The large language models}
An important takeaway of this research is the power of GPT-4. We use the model to reformulate messages to be less inflammatory, and to suggest or directly send interventions to negotiating parties.

The capability of the model to accomplish these tasks is very promising. While a more in-depth evaluation is required to better understand the different failure-models and fully understand the performance, the model seems to be capable of understanding a wide variety of contexts, and providing a relevant and appropriate responses, clearly taking into account the provided instructions and user messages. When reformulating a message, the model appears to maintain the important elements of a message, while at the same time changing the tone. When drafting messages for the mediator, the model has no problem understanding the messages between the parties, and generating a relevant response.

It is interesting how much of these capabilities are emergent behaviour from the model. The prompts we provided are very short, and an interesting approach for future work would be to examine the impact of further ``tuning'' these prompts to achieve better results. However, even so, the model seems to have a remarkable behavior that well corresponds with what it is expected to do. For example, as we can see in Section \ref{sec:message_generation}, the model is essentially only told that it should act as a mediator and encourage the parties to come to a settlement. Based on this simple prompt, the model uses a number of interesting and relevant techniques, such as establishing the facts when they are unclear, reminding the parties to be respectful if necessary, and even suggesting relevant solutions to the parties. It seems likely the model, through being trained on billions of pages of text by OpenAI, has learnt what kind of interventions a mediator may undertake, and even which solutions could be appropriate for different kinds of disputes. 

We want to highlight the fact that all of these tasks were accomplished without any training on our part whatsoever. Previously, a model that is able to read a message and provide a reformulated message, or draft context-specific interventions, is likely to have required a dataset of thousands of existing examples. Even then, it is questionable whether the performance would reach that of the GPT-4 model used in this case, as these are very complex tasks, especially considering the variety of different domains the model seems to be able to handle. 
In a domain such as law, where training data is often difficult to create, finding useful ways to leverage such models may lead to very effective approaches.

At the same time, the models have limitations. For example, they are sometimes prone to hallucination, i.e., making up untruthful information. While this is not something that we witnessed in our experiments, depending on the use case, this may involve certain risks. Further, the texts given by the model may not be accurate. For example, in the final example in table \ref{table:suggestions}, the suggested intervention posits that the seller is responsible for the packaging of the product. This could be seen as giving an opinion, and may not be desirable in a mediation context.

To overcome these potential risks, the LLMediator primarily employs the LLMs to generate suggestions that have to be reviewed by a person before being sent out. Let us discuss the different use-cases for LLMs in the system.

\subsection{F1 - The reformulation of inflammatory messages}
The first LLM-enhanced feature of the LLMediator is the automatic reformulation of messages, to make the messages less inflammatory. This could help maintain the discussion at a positive and neutral tone, and increase the likelihood of the parties finding an amicable settlement. This feature is based on detecting inflammatory messages and then making a request to GPT-4 with the message and a prompt telling the LLM to reformulate it using a more neutral tone.

We discuss two ways this message could then be used by the system in section \ref{sec:show_reformulation}. The first potential way is to automatically reformulate the message, and send the reformulated version to the other party directly. However, this approach has potential negative effects. Any inaccuracies in the reformulated message may lead to misunderstandings, and the sending party may be frustrated that they are not allowed to express themselves freely. Further, it raises interesting questions with regards to self-expression---is it really fair to put words in someones mouth in this manner?

Similar points were discussed in the context of a study that investigated the introduction of a limit on the number of messages that can be sent in a certain time period for individuals discussing controversial topics. Experiment participants did indicate that the limitation made the messages written more intentional. However, they also expressed frustration regarding the imposed limit and its impact on building rapport between the parties \cite{masrani2023slowing}.

Instead, we chose a more feasible and less risky approach, of merely showing the reformulated message to the user, and giving them the choice of which message to send. Here, the AI system acts as an augmented intelligence, that can help the parties convey their meaning in a more neutral way. Seeing a notification by the system telling the user that their message may contain inflammatory language could nudge them towards reconsidering their message. Such an approach has previously been tried by platforms such as twitter, which found that users that were prompted to reconsider harmful or offensive tweet replies were 34\% likely to reformulate their reply, or not send it at all \cite{Butler_Parrella_2021}.

Our approach uses LLMs to additionally suggest a reformulated version to the user. This could be a very powerful way to further encourage them to write their message using more acceptable tone, by making a concrete suggestion, which requires only the push of a button to send. However, the user is always in charge, and thus able to correct any mistakes in the message, or update it to more closely align with what they want to say, while retaining the more neutral tone.

Thus, this system has a lot of potential in supporting the parties to settle their disputes in a friendly and efficient manner. At the same time, as an augmented intelligence system, the risks of hallucinations are relatively low---the system supports the user by suggesting a reformulation, but does not force them to write their message in a certain way. While a deployment of such a system to a real-world ODR platform would, of course, require more empirical evaluation of the risks of such a system, the framing as an augmented intelligence system significantly reduces the risk of such a deployment.

\subsection{F2 - Drafting messages for the mediator}
The second LLM-enhanced feature of the LLMediator is the drafting of a possible intervention for a mediator. In order to utilize this feature, the mediator can initiate a request for a suggestion, which asks the LLM to draft a message based on the previous messages exchanged by the parties. This message is then shown to the mediator, who has the option of editing the suggestion, or sending it to the parties. Further, they have the option to provide additional instructions to the model, which are taken into account by the LLM in drafting a suggestion.

In this task, the model has significantly more freedom than in F1. We are not asking it simply to reformulate a message in a new tone, but to autonomously decide how to respond to the messages of the parties. The responses we examined do this by dynamically adapting to the conversation between the parties, e.g., by calming the discussion, trying to establish key facts, or even suggesting possible compromises that the parties may consider. None of these approaches were part of the prompt---rather, they were chosen by the model itself, when told to mediate the discussion. This kind of emergent capability is very interesting, and an impressive demonstration of the power of GPT-4 to perform different tasks.

At the same time, just like F1, F2 is based on the concept of augmented intelligence. The system does not prescribe to the mediator what they should do, but rather offers a draft of the intervention message. The mediator can use the draft directly, or change it if they do not think it is appropriate. They can also give additional instructions to the model, which will be incorporated in the prompt. As can be seen in Table \ref{table:suggestions}, the model has an impressive ability to take into account both of these instructions and the previous messaging context.

Supporting the mediator in this way could be very important in areas where sufficient numbers of trained mediators or facilitators are not available (compare \cite{branting2022computational}). Verifying that a message is appropriate and helpful could be faster for an experienced mediator than drafting messages from scratch. Thus, fewer mediators could mediate more disputes, which could be an important contribution to access to justice.

At the same time, it is important to be aware of the potential risks of deploying such a system. The interaction between the mediator and the LLM should be closely studied to make sure that the results are not harmful. For example, perhaps a suggested message could anchor a mediator to a certain type of message, even if another approach could have been more helpful. Likewise, the mediator may start to overly rely on the LLM-provided message to understand the previous messages written by the parties, which could be risky if the model starts to hallucinate. Finally, the assumption underlying the AI suggestion feature is that it is faster for a mediator to review and/or edit a suggested message than drafting a new message---empirical studies are needed to confirm whether this is the case.

\subsection{F3 - Autonomously intervening in the negotiation?}
\label{sec:discussion:F3}
The most radical use of the LLMediator is the direct intervention of the system in the negotiation. Here, the system acts as a mediator directly. Just like in F2, a message is drafted. However, instead of sending the message as a suggestion to the mediator, now the messages are sent directly to the parties instead.

This approach carries with it substantial risks---since under this use, no humans verify the message before it goes to the user, any potential biased or inaccurate information it contains may have an effect on the negotiation between the parties. An example of this can be seen in Table \ref{table:suggestions} in the final example, where the LLM seems to indicate that one of the parties was responsible for packaging the product safely. In doing so, it could be seen to take the side of one of the parties, which may not be desirable for a mediator.

Of course, the messages would very clearly indicate that they are written by an AI-based system, and the parties are free to disregard the messages of the mediator---they are, after all, a neutral third party aiming to help the parties, and not an adjudicator. However, even so, biased or inaccurate messages may lead the parties to lose trust in the platform, or even influence the negotiation towards an unfair outcome. Thus, considerable studies would have to be performed before the deployment of such a system.

However, if the model can be prompted and deployed in a manner that is sufficiently safe, this approach could have tremendous implications for access to justice. Every single dispute, no matter how small, could be given a personalized, always available mediator. This could enable a form of mediation to be used in areas where disputes are too small to warrant a mediator intervention, or there are simply not enough mediators available. If these types of interventions could be shown to increase the rate of settlement between parties, this could be a significant contribution to society, by potentially creating a new way for individuals to resolve their legal problems.

\section{Future Work}
This work lays the groundwork for a number of important research directions.

First, an important avenue is the empirical evaluation of the presented system. Understanding the actual efficacy of the system in dealing with conflicts is crucial in understanding whether the generated reformulations and interventions are relevant and useful, how they may be adapted, and how such a system may eventually be deployed. Such an evaluation might be done with experienced mediators who could assess the quality of the messages, and students taking the roles of parties in a negotiation. The evaluation should also examine whether the system presents any biases.

Another avenue to explore is the further improvement and adaptation of the system. For example, while the prompts appear to work well, so-called ``prompt engineering'' may be an important step towards further improving the results, by altering the prompt sent to the LLM. There are also multiple features described in this paper that are not yet fully developed, such as the function to trigger a reformulation suggestion (see section \ref{sec:detecting}) and the function to trigger an intervention by the AI-mediator (see \ref{sec:triggers}). These are important features which need to be explored and elaborated.

Finally, there may be other interesting ways of integrating LLMs into an ODR platform. For example, such a system could summarize the conversation held between the parties for the mediator or adjudicator. Further building on the LLMediator to explore such use-cases will be an important avenue for future work.

\section{Conclusion}
We presented the LLMediator, an experimental platform built to support negotiation and mediation using LLMs, such as GPT-4. Currently, the system is able to reformulate inflammatory messages, and suggest possible interventions to the mediator, or even send them directly to the parties. We show a few examples of the system being used in fictitious disputes, and discuss its efficacy. Overall, the approach seems promising and opens up the door for a lot of further research.

\begin{acknowledgments}
    Hannes Westermann and Karim Benyeklef would like to thank the Cyberjustice Laboratory at Université de Montréal, the LexUM Chair on Legal Information and the Autonomy through Cyberjustice Technologies (ACT) project for their support of this research.
\end{acknowledgments}

\bibliography{sample-ceur}

\begin{thebibliography}{44}
\bibitem[{Cole and Blankley(2006)}]{cole2006online}
\bibinfo{author}{S.~R. Cole}, \bibinfo{author}{K.~M. Blankley},
\newblock \bibinfo{title}{Online mediation: Where we have been, where we are
  now, and where we should be},
\newblock \bibinfo{journal}{U. Tol. L. Rev.} \bibinfo{volume}{38}
  (\bibinfo{year}{2006}) \bibinfo{pages}{193}.
\end{thebibliography}

\end{document}